\documentclass[final,5p,times,twocolumn,authoryear]{elsarticle}

\usepackage{amssymb}
\usepackage{lipsum}
\usepackage{amsmath}

\begin{document}

\begin{frontmatter}

\title{Robust Cross-Modal Foundation Model Perception for Underwater Robots under Degraded Visual Conditions}

\author[first]{Mohammad Arif Ul Alam}
\affiliation[first]{organization={College of Science and Technology, North Carolina A \& T State University},
            addressline={}, 
            city={Lowell},
            postcode={}, 
            state={MA},
            country={USA}}
\begin{abstract}
Reliable underwater robotic perception remains challenging because optical imagery can deteriorate substantially under turbidity, wavelength-dependent attenuation, low illumination, scattering, and blur. Although sonar provides complementary information that is less sensitive to optical visibility, existing visual-sonar studies have primarily emphasized multimodal feature alignment and nominal detection performance. In this work, we investigate \emph{cross-modal robustness under changing visual reliability} and examine whether a pretrained visual foundation model can provide robust representations while complementary sonar information compensates when optical evidence becomes severely degraded. Specifically, we employ frozen DINOv2 as the visual foundation-model encoder and construct a controlled five-level degradation benchmark ranging from clean to extreme visual conditions. We evaluate conventional visual detection, frozen foundation-model representations, sonar context, fixed multimodal fusion, clean-trained adaptive gating, and degradation-aware gated fusion. Our degradation-aware strategy explicitly exposes the fusion mechanism to the complete range of visual reliability conditions while keeping the foundation-model and sonar encoders frozen, enabling modality contributions to adapt without task-specific fine-tuning of the pretrained backbone. Under the most severe combined degradation, the DINOv2 visual foundation-model baseline achieves a balanced accuracy of $0.4610$, whereas degradation-aware visual-sonar fusion reaches $0.6152$, corresponding to a $33.5\%$ relative improvement. Concurrently, the learned sonar contribution increases from $14.2\%$ under clean conditions to $41.3\%$ under extreme degradation, indicating a systematic redistribution of cross-modal reliance as visual evidence deteriorates. Degradation-specific analysis further reveals the largest fusion benefits under severe turbidity and blur, while color attenuation alone provides little additional gain. These findings indicate that pretrained foundation-model representations provide substantial visual robustness but remain insufficient under severe information loss, and that robust underwater multimodal perception benefits from explicitly training the fusion mechanism to adapt to changes in modality reliability. This preliminary study provides a foundation for future end-to-end, degradation-aware visual-sonar perception systems for underwater robots.
\end{abstract}

\begin{keyword}
Underwater robotics, multimodal perception, visual–sonar fusion, foundation models, degradation-aware fusion.
\end{keyword}

\end{frontmatter}

\section{Introduction}

\label{sec:underwater_degradation}

Reliable visual perception remains challenging for underwater robots because the propagation of light through water is strongly affected by absorption and scattering \cite{Zhu}. As visibility deteriorates, optical imagery can exhibit wavelength-dependent color attenuation, reduced illumination, loss of contrast, scattering-induced veiling, and blur \cite{akkaynak2018revised, akkaynak2019seathru,li2020diving}. These effects can substantially alter the visual evidence available to perception models and create a distribution shift between favorable training imagery and more difficult deployment conditions \cite{Grimaldi2023InvestigationOT}. Sonar provides a complementary sensing modality because acoustic measurements are not directly affected by optical visibility \cite{Sorensen2023Commercial}. However, sonar generally provides less appearance and texture information than optical imagery and introduces its own sensing limitations\cite{SHEN2025121929}. Consequently, optical and sonar sensing are better viewed as complementary rather than interchangeable modalities \cite{wu2026multimodal}. This motivates a central question for underwater autonomy: how should a perception system respond when the reliability of its visual modality progressively decreases?

Existing visual-sonar research has primarily focused on learning effective representations across heterogeneous sensing modalities \cite{HE2026126793, Qiao2026Lightweight, 11397899, Fitzpatrick2023Multimodal, Zhu}. On UMOD, for example, VSAFDet addresses visual-sonar feature heterogeneity and spatial misalignment through specialized cross-modal feature interaction \cite{wu2026multimodal}. Importantly, its experiments also show that simple multimodal addition or concatenation does not necessarily improve performance, indicating that the availability of complementary sensors alone is insufficient for effective fusion. In this work, we study a related but different problem: \emph{multimodal robustness under changing sensor reliability}. Rather than evaluating fusion only under nominal conditions, we progressively degrade the visual observations while leaving the paired sonar measurements unchanged. This allows us to examine whether multimodal perception remains useful as optical reliability deteriorates and whether the fusion mechanism learns to increase its reliance on acoustic information when visual evidence becomes unreliable.

We use DINOv2 as a frozen visual foundation-model encoder because large-scale self-supervised representations provide a useful means of studying whether general visual features remain discriminative under underwater degradation \cite{oquab2023dinov2}. Rather than fine-tuning the foundation model, we keep its parameters fixed and evaluate how its representations change across controlled degradation levels. We then combine the visual representation with paired sonar context using lightweight fusion mechanisms. In particular, we compare fixed fusion, gating trained only under clean conditions, and \emph{degradation-aware gating} trained across the full degradation range. The latter explicitly exposes the fusion mechanism to changes in visual reliability, allowing us to investigate whether acoustic contribution increases as optical conditions deteriorate.

In this paper, we investigate three questions: (i) how visual perception deteriorates under progressive underwater degradation, (ii) whether sonar context can preserve perception when visual information becomes unreliable, and (iii) whether explicit degradation-aware training enables the fusion mechanism to adapt its modality reliance.

The main contributions are:
\begin{itemize}
    \item A controlled five-level ($D_0$-$D_4$) benchmark incorporating illumination loss, wavelength-dependent attenuation, scattering, and blur;
    
    \item An empirical comparison of conventional visual detection, frozen DINOv2 representations, sonar context, and lightweight visual-sonar fusion under progressive degradation;
    
    \item A degradation-aware gated fusion strategy that learns to redistribute visual and sonar contributions as optical reliability changes; and
    
    \item An analysis of modality reliance and degradation type, showing that acoustic assistance is particularly beneficial under severe turbidity and blur.
\end{itemize}

The experiments show that degradation-aware fusion preserves substantially more recognition performance under severe visual degradation than visual-only or nominally trained fusion models. These results provide preliminary evidence that robust underwater multimodal perception depends not only on combining complementary sensors, but also on exposing the fusion mechanism to changes in sensor reliability during training.

\section{Related Work}
\label{sec:related_work}

\subsection{Underwater Object Perception}

Underwater optical perception is commonly addressed through image enhancement and task-specific detection architectures designed to mitigate low contrast, color distortion, blur, and small-object appearance. Recent deep detectors increasingly employ multi-scale feature aggregation, attention, and specialized convolutional modules to improve recognition in challenging underwater scenes \cite{wang2022underwater,zhang2024lightweight}. Nevertheless, optical sensing remains intrinsically sensitive to visibility and illumination. Acoustic sensing provides an alternative when optical observations become unreliable, and learning-based sonar methods have demonstrated improved extraction of structural and geometric target information from noisy acoustic imagery \cite{yang2024sonar,wen2024sonar}. These complementary characteristics motivate joint optical-acoustic perception rather than reliance on either modality alone.

\subsection{Visual-Sonar Multimodal Fusion}

Multimodal underwater perception has increasingly explored visual-sonar fusion for target recognition and detection. Prior studies have combined optical imagery with synthetic-aperture or side-scan sonar for underwater target classification, demonstrating that acoustic measurements can provide complementary information under poor visibility \cite{abu2023multimodal,wang2023multimodal}. More recent detection architectures perform feature-level fusion between synchronized visual and sonar streams. UAMFDet, for example, employs adaptive multimodal feature alignment for underwater detection \cite{chen2025uamfdet}. A persistent difficulty is that visual and sonar measurements differ substantially in resolution, imaging geometry, and information density, making direct feature correspondence unreliable. Attention-based and cross-modal interaction mechanisms have therefore become important alternatives to simple feature aggregation. Notably, experiments on UMOD show that direct addition or concatenation can introduce inter-modal interference rather than improve detection, further motivating adaptive fusion mechanisms.

\subsection{Robust and Adaptive Multimodal Perception}

Robust multimodal learning requires more than maximizing nominal fusion performance. When one modality becomes corrupted or unreliable, a fixed fusion strategy can propagate degraded features into the joint representation. Research on multimodal robustness has therefore considered modality dropout, missing-modality learning, and reliability-aware fusion to reduce dependence on consistently available high-quality inputs \cite{neverova2016moddrop,ma2022smil}. Related work in multimodal autonomous perception has also shown that complementary sensing can improve robustness under adverse environmental conditions when cross-modal interactions are explicitly modeled \cite{wu2023robust}. However, underwater visual-sonar research has largely emphasized feature alignment and detection accuracy rather than systematically measuring how learned modality reliance changes as optical quality progressively deteriorates. This work addresses that distinction by explicitly exposing the fusion model to multiple degradation severities and evaluating whether it learns to shift reliance toward sonar as visual evidence becomes less reliable.

\subsection{Foundation Models under Distribution Shift}

Large pretrained visual models provide transferable representations that can support downstream tasks with limited task-specific supervision. Recent studies have examined the robustness of self-supervised and vision-transformer representations under common corruptions and distribution shifts, often finding stronger representation stability than conventional supervised models \cite{hendrycks2021pretraining,paul2022vision}. Such robustness is particularly relevant to underwater perception, where deployment conditions can differ substantially from terrestrial pretraining data. Nevertheless, foundation-model representations alone cannot recover information that is physically lost under severe scattering, blur, or visibility reduction. This motivates our use of a frozen foundation-model representation together with complementary sonar context, while placing the primary emphasis on \emph{degradation-aware cross-modal adaptation} rather than foundation-model fine-tuning.

\section{Problem Formulation}
\label{sec:problem_formulation}

We formulate underwater cross-modal perception as a recognition problem in which an optical observation may undergo a progressive loss of reliability while its temporally paired sonar observation remains unchanged. The objective is not merely to determine whether visual and acoustic information are complementary, but to characterize how perception performance changes as the quality of one modality deteriorates and whether a multimodal model can adapt its reliance on the remaining information. This formulation separates \emph{modality availability} from \emph{modality reliability}: both modalities remain present throughout the experiment, but the information carried by the optical modality is systematically degraded.

\subsection{Cross-Modal Underwater Observation}
\label{sec:crossmodal_observation}

Let a synchronized visual-sonar observation be represented as
\begin{equation}
    \mathbf{x} = \left(\mathbf{x}_v,\mathbf{x}_s\right),
\end{equation}
where $\mathbf{x}_v \in \mathcal{X}_v$ denotes an optical image and $\mathbf{x}_s \in \mathcal{X}_s$ denotes its temporally synchronized sonar observation. Each annotated visual object is associated with a class label
\begin{equation}
    y \in \mathcal{Y}=\{1,\ldots,C\},
\end{equation}
where $C$ is the number of target classes retained for evaluation.

The two observations should not be interpreted as pixel-wise registered views of the same scene. Optical cameras form perspective images in an image plane, whereas forward-looking sonar represents acoustic returns in a range-azimuth geometry. Consequently, corresponding visual and sonar observations can exhibit substantial differences in spatial coordinates, resolution, appearance, and information density. The UMOD acquisition system nevertheless provides synchronized observations from the two sensors, making it possible to study their complementary information without assuming direct pixel-level correspondence \cite{wu2026multimodal}.

For the recognition experiments considered in this study, the visual input associated with an annotated object is denoted by
\begin{equation}
    \mathbf{x}_v^{(o)} = \mathcal{C}(\mathbf{x}_v,b),
\end{equation}
where $b$ is the ground-truth visual bounding box and $\mathcal{C}(\cdot)$ denotes the object-cropping operation. The corresponding sonar input is the synchronized sonar frame $\mathbf{x}_s$. Thus, the multimodal prediction problem is
\begin{equation}
    \hat{y}
    =
    F\!\left(\mathbf{x}_v^{(o)},\mathbf{x}_s\right),
\end{equation}
where $F(\cdot)$ denotes a visual-sonar recognition model. This formulation deliberately treats sonar as \emph{paired acoustic context} rather than as a geometrically registered object crop. It therefore allows us to isolate whether synchronized acoustic information can compensate for a loss of discriminative visual evidence without requiring an explicit visual-to-sonar coordinate transformation.

\subsection{Controlled Visual Degradation Model}
\label{sec:degradation_model}

To study robustness under changing optical reliability, degradation is applied only to the visual observation. Let
\begin{equation}
    \tilde{\mathbf{x}}_v^{(s)}
    =
    \mathcal{D}\!\left(\mathbf{x}_v;s\right),
    \qquad
    s\in\{0,1,2,3,4\},
\end{equation}
where $\mathcal{D}(\cdot;s)$ is a degradation operator and $s$ denotes its severity. We define five ordered operating conditions,
\begin{equation}
    D_0,D_1,D_2,D_3,D_4,
\end{equation}
corresponding respectively to clean, mild, moderate, severe, and extreme visual degradation. The sonar observation is held fixed:
\begin{equation}
    \tilde{\mathbf{x}}_s^{(s)}=\mathbf{x}_s
    \qquad \forall s.
\end{equation}
This construction creates a controlled reliability shift in which increasing $s$ reduces the quality of the optical evidence while preserving the paired acoustic observation.

\begin{figure*}[!htb]
    \centering
    \includegraphics[width=0.95\textwidth]{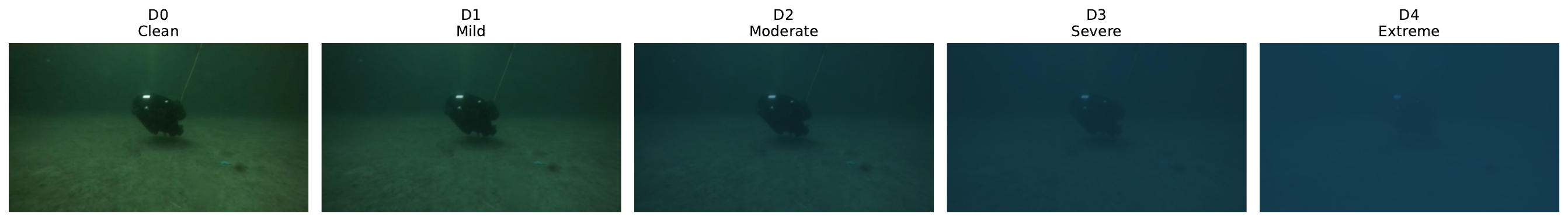}
    \caption{Illustration of the controlled visual degradation sequence used in the robustness benchmark. From left to right, the optical observation progresses from $D_0$ (clean) through $D_1$ (mild), $D_2$ (moderate), $D_3$ (severe), and $D_4$ (extreme) degradation. The paired sonar observation is held unchanged across all degradation levels.}
    \label{fig:degradation_levels}
\end{figure*}

The combined degradation operator models four common manifestations of adverse underwater optical conditions: illumination loss, wavelength-dependent attenuation, scattering or turbidity, and blur. These transformations are intended as controlled perturbations for robustness evaluation rather than as a complete physical simulation of underwater image formation.

\subsubsection{Illumination Attenuation}

Reduction in available illumination is represented by an intensity-scaling operation
\begin{equation}
    I_{\mathrm{ill}}(\mathbf{p})
    =
    \alpha_s I(\mathbf{p}),
    \qquad 0<\alpha_s\leq 1,
\end{equation}
where $I(\mathbf{p})$ is the original pixel intensity at location $\mathbf{p}$ and $\alpha_s$ decreases with degradation severity. This transformation progressively suppresses visual contrast and low-intensity details.

\subsubsection{Wavelength-Dependent Attenuation}

Light attenuation underwater is wavelength dependent; in particular, longer wavelengths such as red are generally attenuated more strongly than shorter blue-green wavelengths. We approximate this effect by applying severity-dependent channel coefficients,
\begin{equation}
    I_{\mathrm{att}}^{c}(\mathbf{p})
    =
    \beta_{c,s} I^{c}(\mathbf{p}),
    \qquad
    c\in\{R,G,B\},
\end{equation}
with
\begin{equation}
    \beta_{R,s} < \beta_{G,s} < \beta_{B,s}
\end{equation}
for degraded conditions. The coefficients become progressively stronger with increasing $s$, producing increasing color distortion and loss of wavelength-dependent visual information.

\subsubsection{Turbidity and Scattering}

Scattering and suspended particles reduce scene contrast while introducing veiling light. We approximate this behavior using the standard attenuation-backscatter form
\begin{equation}
    I_{\mathrm{turb}}(\mathbf{p})
    =
    J(\mathbf{p})t_s
    +
    A_s(1-t_s),
\end{equation}
where $J(\mathbf{p})$ represents the undegraded scene radiance, $t_s\in[0,1]$ is a severity-dependent transmission coefficient, and $A_s$ represents ambient or veiling illumination. As degradation becomes stronger, $t_s$ decreases, causing the observed image to contain progressively less scene information and a larger scattering component.

\subsubsection{Blur}

Loss of fine spatial structure is represented using Gaussian smoothing,
\begin{equation}
    I_{\mathrm{blur}}
    =
    G_{\sigma_s} * I,
\end{equation}
where $G_{\sigma_s}$ is a Gaussian kernel with standard deviation $\sigma_s$, $*$ denotes convolution, and $\sigma_s$ increases with severity. The operation suppresses high-frequency information such as object edges, texture, and fine structural details.

The combined benchmark applies severity-dependent instances of these transformations to produce the sequence $D_0$-$D_4$. Figure~\ref{fig:degradation_levels} provides a representative example of the resulting degradation sequence. The progression is intended to create an ordered loss of useful optical evidence rather than to map each level to a specific physical visibility distance or water type. This formulation does not claim that a single scalar severity corresponds to a specific physical water condition or visibility distance. Instead, the ordered levels provide a reproducible means of evaluating how rapidly different perception strategies lose performance as optical evidence is progressively corrupted.

\subsection{Robustness Objective}
\label{sec:robustness_objective}

Let $F_m$ denote a perception strategy $m$ and let $M_m(s)$ denote its evaluation score under degradation level $D_s$. For visual-only perception,
\begin{equation}
    \hat{y}_{v}^{(s)}
    =
    F_v\!\left(\tilde{\mathbf{x}}_v^{(s)}\right),
\end{equation}
whereas cross-modal perception is represented by
\begin{equation}
    \hat{y}_{vs}^{(s)}
    =
    F_{vs}\!\left(
        \tilde{\mathbf{x}}_v^{(s)},
        \mathbf{x}_s
    \right).
\end{equation}

Absolute performance alone does not distinguish clean-condition capability from degradation sensitivity. We therefore define the retained performance of model $m$ at severity $s$ relative to its clean-condition score as
\begin{equation}
    R_m(s)
    =
    \frac{M_m(s)}{M_m(0)},
    \qquad s\in\{1,2,3,4\}.
\end{equation}
A value $R_m(s)=1$ indicates that the measured performance is unchanged relative to $D_0$, while values below one indicate degradation relative to the clean condition.

To summarize behavior over the complete degradation range, we define the mean relative robustness (MRR) as
\begin{equation}
    \mathrm{MRR}_m
    =
    \frac{1}{4}
    \sum_{s=1}^{4}
    R_m(s)
    =
    \frac{1}{4}
    \sum_{s=1}^{4}
    \frac{M_m(s)}{M_m(0)}.
\end{equation}
MRR measures average performance retention rather than absolute predictive accuracy and is therefore interpreted together with the underlying task metric. In particular, an MRR close to or slightly greater than one should be interpreted as approximate preservation of clean-condition performance across the tested perturbations, rather than evidence that visual degradation itself improves perception.

The central robustness objective is consequently to learn a cross-modal predictor whose performance degrades more slowly than its visual-only counterpart:
\begin{equation}
    R_{vs}(s) > R_v(s)
    \quad
    \text{as } s \text{ increases},
\end{equation}
particularly under severe conditions $D_3$ and $D_4$. More specifically, we seek a fusion function whose use of the two modalities is conditional on their available evidence rather than fixed across operating conditions. If $g_v^{(s)}$ and $g_s^{(s)}$ denote the learned visual and sonar contributions, respectively, with
\begin{equation}
    g_v^{(s)}+g_s^{(s)}=1,
\end{equation}
the desired qualitative behavior under increasing visual degradation is
\begin{equation}
    g_s^{(s)} \uparrow
\quad \text{as visual evidence becomes less informative},
\end{equation}
when the acoustic representation provides useful complementary evidence. The expected aggregate behavior is an increased contribution from sonar under severe visual degradation; strict monotonicity across adjacent severity levels is neither imposed nor required. Importantly, these coefficients are treated as learned fusion weights rather than calibrated probabilities of sensor reliability.

This formulation leads to the principal hypothesis examined in the remainder of the paper: \emph{robust cross-modal perception requires not only complementary modalities, but also a fusion mechanism exposed during training to changes in modality reliability}. The subsequent experiments therefore distinguish between fixed fusion, adaptive fusion learned only from clean observations, and degradation-aware adaptive fusion learned across $D_0$-$D_4$.

\section{Dataset and Experimental Protocol}
\label{sec:experimental_protocol}

\subsection{Dataset and Problem Setup}
\label{sec:dataset_setup}

\begin{figure*}[!htb]
    \centering
    \includegraphics[width=0.7\textwidth]{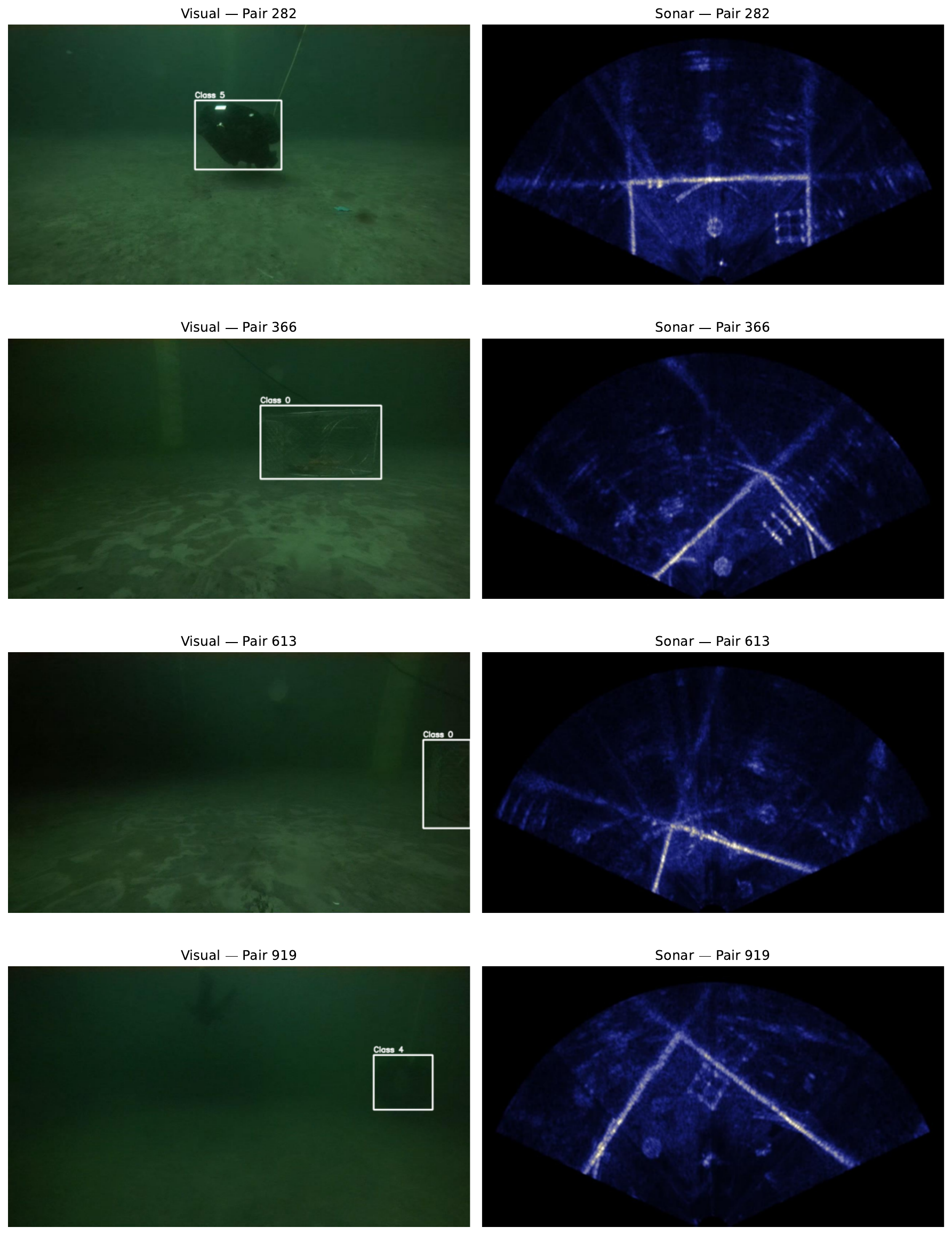}
    \caption{Representative synchronized visual-sonar observations from the UMOD subset used in this study. The optical images contain annotated visual objects, while the paired forward-looking sonar frames provide acoustic scene context. The modalities are temporally synchronized but are not assumed to be pixel-wise or geometrically registered.}
    \label{fig:umod_pairs}
\end{figure*}

We conduct our experiments using the Underwater Multimodal Object
Detection (UMOD) dataset introduced by Wu et al.~\cite{wu2026multimodal}.
UMOD was designed specifically for studying multimodal underwater
perception using temporally synchronized optical and acoustic
observations. The data were collected using a remotely operated vehicle
(ROV) in a $7\,\mathrm{m}\times6\,\mathrm{m}\times5\,\mathrm{m}$ testing
pool at the Sanya Ocean Research Institute in Hainan Province, China.
The sensing platform combines a BlueView M900 two-dimensional
forward-looking sonar with a ZED2 underwater stereo optical camera. The
sonar provides a $130^{\circ}$ field of view, a reported resolution of
$6.25\,\mathrm{cm}$, and a maximum sensing range of $100\,\mathrm{m}$,
while the optical camera supports image resolutions of
$1280\times800$ and $640\times400$ at up to $60$ frames per second.
Optical and sonar observations were acquired simultaneously using
synchronized timestamps, and frames were extracted from the
synchronized streams at five-frame intervals to construct paired
multimodal observations \cite{wu2026multimodal}. Representative synchronized observations are shown in Fig.~\ref{fig:umod_pairs}. The examples illustrate both the complementarity and the substantial representational difference between the two sensing modalities. Optical imagery provides comparatively rich appearance information, whereas forward-looking sonar represents the scene through acoustic returns in a distinct range-azimuth geometry. Accordingly, the paired observations are treated as temporally corresponding but not pixel-wise registered.

After removal of redundant observations and the preprocessing procedure
described by the dataset authors, UMOD contains $4{,}000$ synchronized
visual-sonar image pairs, corresponding to $8{,}000$ modality-specific
images. The dataset contains nine stationary and moving underwater
target categories:
\emph{cage}, \emph{frame}, \emph{hook}, \emph{anchor}, \emph{tire},
\emph{ROV}, \emph{plastic bucket}, \emph{fish}, and \emph{oil drums}.
The original UMOD benchmark was divided by its authors into training,
validation, and test sets using a $7{:}2{:}1$ ratio
\cite{wu2026multimodal}. In addition to providing temporally
corresponding observations, the dataset captures the substantial
heterogeneity between the two sensing modalities: optical imagery
contains comparatively rich appearance, color, and texture information,
whereas forward-looking sonar primarily represents acoustic structure
in range-azimuth coordinates. UMOD is therefore particularly suitable
for studying whether acoustic information can complement optical
representations as visual reliability deteriorates.

Our experiments formulate the problem at two related levels. First, we
use visual object detection to quantify the effect of progressive
degradation on a conventional detector. Second, the primary
cross-modal experiments formulate object recognition using the
ground-truth visual object crop and its temporally paired sonar frame,
as defined in Section~\ref{sec:crossmodal_observation}. This separation
allows detection-level degradation sensitivity to be measured while
isolating the behavior of visual foundation-model representations and
visual-sonar fusion from errors introduced by object localization.

For the recognition and fusion experiments, we retain five target
categories corresponding to the original UMOD class identifiers
$\{0,1,3,4,5\}$. These classes are remapped to contiguous labels
$\{0,\ldots,4\}$ for model training and evaluation. The same class
definition is maintained across the visual-only, sonar-context, fixed
fusion, and adaptive-fusion experiments so that changes in performance
reflect differences in representation and modality use rather than
changes in the prediction task.

Because UMOD observations are extracted from synchronized video
sequences, randomly assigning individual frames to different data
partitions can place temporally adjacent and visually similar
observations in both training and evaluation sets. Such a split can
inflate measured generalization performance by introducing
sequence-level leakage. We therefore construct a fixed grouped split
before model training. Contiguous sample identifiers are partitioned
into non-overlapping blocks of $40$ identifiers, and entire blocks,
rather than individual observations, are assigned to the training,
validation, or test partition. Block assignment is performed while
considering the retained class distribution so that all evaluated
classes remain represented as consistently as possible across
partitions. Once generated, the split is locked and reused for every
model and every degradation condition.

This protocol is especially important for the degradation experiments.
All transformed versions of a given clean visual observation remain in
the same partition as their corresponding $D_0$ observation. Thus, if
$\mathbf{x}_v$ belongs to the test set, each
$\mathcal{D}(\mathbf{x}_v;s)$ for $s\in\{0,\ldots,4\}$ also belongs
exclusively to the test set. Likewise, the synchronized sonar
observation $\mathbf{x}_s$ never crosses partition boundaries.
Consequently, the degradation benchmark changes only the reliability
of the visual evidence while preserving sample identity, class label,
sonar context, and train-validation-test membership. This enables
paired comparisons across $D_0$-$D_4$ and prevents degraded variants
of an evaluation observation from being encountered during training.

\subsection{Evaluation Metrics}
\label{sec:evaluation_metrics}

The experiments include both object detection and object recognition,
and we therefore report metrics appropriate to each task. For the
visual detection experiments, performance is evaluated using mean
average precision (mAP), consistent with standard underwater object
detection evaluation and the original UMOD study
\cite{wu2026multimodal}. For a class $c$, average precision is computed
from its precision-recall curve as

\begin{equation}
    \mathrm{AP}_c
    =
    \int_{0}^{1} P_c(r)\,dr,
\end{equation}

where $P_c(r)$ denotes precision as a function of recall. Mean average
precision over $C$ evaluated classes is

\begin{equation}
    \mathrm{mAP}
    =
    \frac{1}{C}
    \sum_{c=1}^{C}
    \mathrm{AP}_c.
\end{equation}

We report $\mathrm{mAP}@0.5$, in which a predicted bounding box is
matched to a ground-truth object using an intersection-over-union (IoU)
threshold of $0.5$, together with the more stringent
$\mathrm{mAP}@0.5{:}0.95$, which averages AP over IoU thresholds from
$0.50$ to $0.95$ in increments of $0.05$. The former emphasizes
successful object detection under a moderate localization criterion,
whereas the latter is more sensitive to localization quality across a
range of overlap requirements. Precision and recall are additionally
used where useful for interpreting detector behavior.

For the object-level foundation-model and cross-modal recognition
experiments, we report classification accuracy and balanced accuracy.
For $N$ evaluation objects, classification accuracy is

\begin{equation}
    \mathrm{Acc}
    =
    \frac{1}{N}
    \sum_{i=1}^{N}
    \mathbb{I}\!\left(\hat{y}_i=y_i\right),
\end{equation}

where $\mathbb{I}(\cdot)$ is the indicator function. Because the
frequency of target categories is not necessarily uniform, overall
accuracy can be dominated by more frequent classes. We therefore use
balanced accuracy as the primary recognition metric:

\begin{equation}
    \mathrm{BAcc}
    =
    \frac{1}{C}
    \sum_{c=1}^{C}
    \frac{\mathrm{TP}_c}
         {\mathrm{TP}_c+\mathrm{FN}_c},
\end{equation}

where $\mathrm{TP}_c$ and $\mathrm{FN}_c$ denote the true-positive and
false-negative counts for class $c$, respectively. Balanced accuracy
assigns equal importance to the recall of each target category and is
therefore better suited to comparisons in which the class distribution
is imbalanced.

Robustness is evaluated by measuring each model under the complete
ordered degradation sequence $D_0$-$D_4$. In addition to the absolute
task metric at each severity, we report retained performance relative
to the corresponding clean condition using $R_m(s)$ and summarize
performance retention over the degraded conditions using the mean
relative robustness (MRR) defined in
Section~\ref{sec:robustness_objective}. For recognition experiments,
$M_m(s)$ is instantiated using balanced accuracy, whereas for the
visual detector it is instantiated using the corresponding mAP metric.
This distinction is necessary because detection and object-level
recognition constitute different prediction tasks; their absolute
scores are therefore not interpreted as directly comparable measures
of model quality. Instead, comparisons across these tasks emphasize
the rate at which each perception strategy loses performance relative
to its own clean-condition baseline.

Finally, all models are evaluated on identical sample partitions and,
within each task, on identical degradation realizations. The clean
condition $D_0$ serves as the reference operating point, while
$D_1$-$D_4$ represent progressively stronger perturbations. This
paired evaluation design allows changes in performance to be attributed
to controlled changes in visual reliability rather than differences in
test composition. Together, absolute task performance, balanced
accuracy, retained performance, and MRR characterize both predictive
capability and sensitivity to progressive visual degradation.

\section{Method}
\label{sec:method}

\begin{figure*}[!htb]
    \centering
    \includegraphics[width=0.92\textwidth]{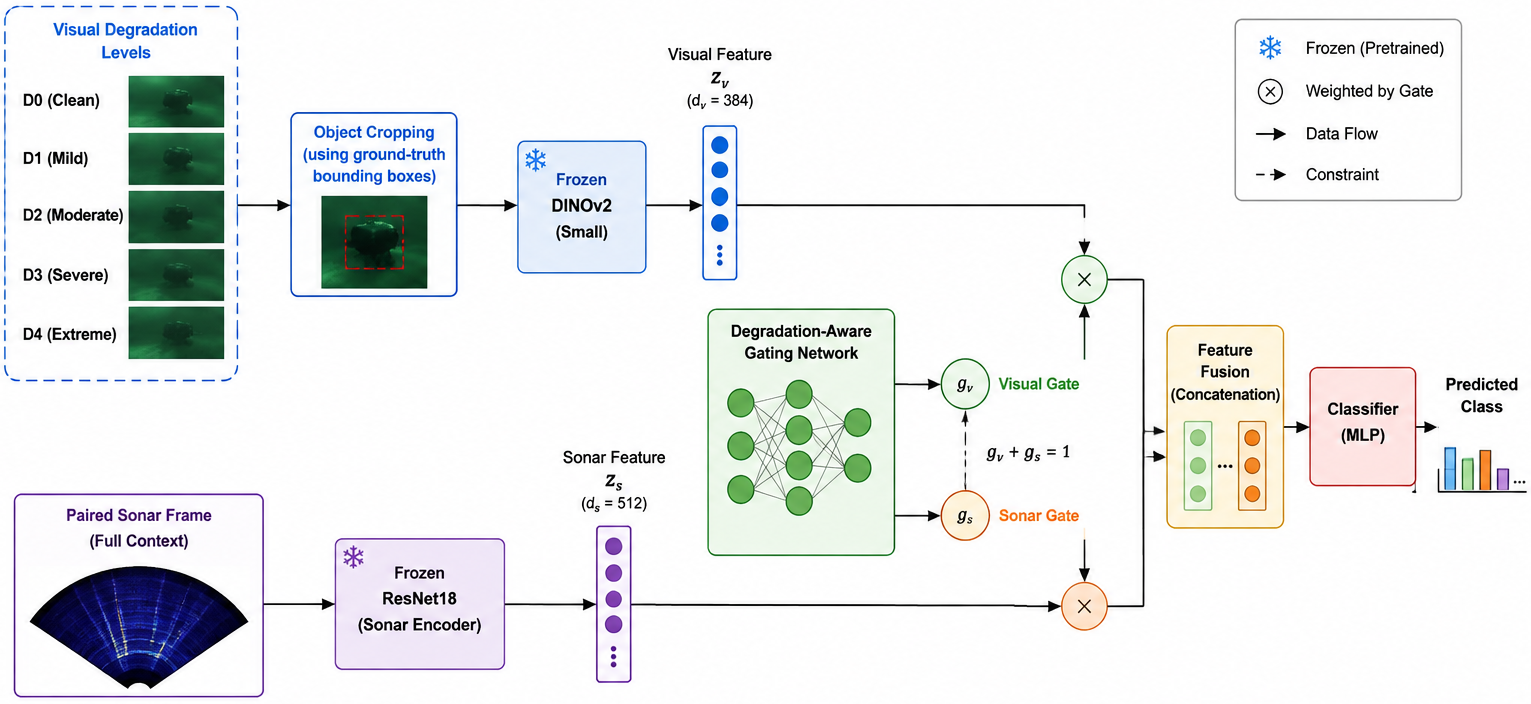}
    \caption{Overview of the proposed degradation-aware cross-modal perception framework. The visual object crop is progressively degraded from $D_0$ to $D_4$ and encoded using a frozen DINOv2 backbone, while the synchronized sonar frame is represented by a frozen ResNet-18 encoder. The projected modality representations are combined through learned adaptive gating, allowing their relative contributions to vary with the available visual evidence.}
    \label{fig:method_overview}
\end{figure*}
The proposed framework is designed to isolate the effect of changing visual reliability on cross-modal underwater perception. Rather than constructing an end-to-end visual-sonar detector, we decouple representation learning from multimodal fusion. An annotated optical object is first represented using a frozen visual foundation model, while its synchronized sonar frame is encoded independently as acoustic context. The resulting modality-specific representations are then combined using either fixed feature fusion or a lightweight adaptive gating mechanism. This design is motivated by the substantial heterogeneity and spatial misalignment between optical and forward-looking sonar observations in UMOD \cite{wu2026multimodal}, while allowing the fusion process to be studied independently of object-localization errors.

Let $\tilde{\mathbf{x}}_v^{(s,o)}$ denote the visual crop associated with an annotated object after applying degradation level $D_s$, and let $\mathbf{x}_s$ denote its synchronized sonar frame. The complete cross-modal recognition pipeline can be written as
\begin{equation}
    \mathbf{z}_v^{(s)}
    =
    f_v\!\left(\tilde{\mathbf{x}}_v^{(s,o)}\right),
    \qquad
    \mathbf{z}_s
    =
    f_s\!\left(\mathbf{x}_s\right),
\end{equation}
followed by
\begin{equation}
    \hat{y}^{(s)}
    =
    h\!\left(
        \Phi\!\left(\mathbf{z}_v^{(s)},\mathbf{z}_s\right)
    \right),
\end{equation}
where $f_v(\cdot)$ and $f_s(\cdot)$ are the visual and sonar encoders, respectively, $\Phi(\cdot)$ denotes the fusion operation, and $h(\cdot)$ is the downstream classifier. Only the optical observation varies across $D_0$-$D_4$; the corresponding sonar representation remains unchanged for a given synchronized sample. This construction enables changes in the learned fusion behavior to be attributed specifically to changes in visual evidence.

\subsection{Visual Foundation-Model Representation}
\label{sec:visual_representation}

We use DINOv2 as the visual representation backbone. DINOv2 is a self-supervised visual foundation model trained to produce transferable representations without requiring task-specific labels during pretraining \cite{oquab2023dinov2}. Its use in this study serves two purposes. First, it provides a strong pretrained representation that is not learned specifically from the relatively small underwater subset used in our experiments. Second, freezing the encoder permits the robustness of the representation itself to be evaluated as visual quality changes, without confounding this behavior with degradation-specific fine-tuning.

For an object with ground-truth visual bounding box $b$, the corresponding crop is extracted from the optical image with a small amount of surrounding context. In our implementation, the bounding box is expanded by $10\%$ before cropping, subject to the image boundaries. The resulting object image is resized and normalized according to the DINOv2 input preprocessing procedure and passed through a frozen DINOv2-small encoder. The visual representation is therefore
\begin{equation}
    \mathbf{z}_v^{(s)}
    =
    f_{\mathrm{DINOv2}}
    \left(
        \mathcal{C}
        \left(
            \tilde{\mathbf{x}}_v^{(s)},b
        \right)
    \right),
\end{equation}
where $\mathcal{C}(\cdot)$ denotes the padded cropping operation. The resulting embedding is
\begin{equation}
    \mathbf{z}_v^{(s)}
    \in
    \mathbb{R}^{384}.
\end{equation}

The DINOv2 parameters remain fixed throughout all recognition and fusion experiments. Consequently, the same visual encoder is used for clean and degraded observations, and no degradation-specific adaptation occurs inside the foundation model. This is important to the experimental design: any variation in $\mathbf{z}_v^{(s)}$ arises from the changing optical input rather than from changes to the encoder parameters.

A lightweight classifier is trained on top of these frozen embeddings to establish the visual foundation-model baseline. Let $h_v(\cdot)$ denote this classifier. Visual-only prediction is given by
\begin{equation}
    \hat{y}_v^{(s)}
    =
    h_v\!\left(\mathbf{z}_v^{(s)}\right).
\end{equation}
The same representation is subsequently used by the multimodal models, ensuring that differences between visual-only and fused recognition originate from the incorporation of sonar information and the fusion strategy rather than from different visual backbones.

\subsection{Sonar Context Representation}
\label{sec:sonar_representation}

The acoustic branch is intentionally constructed as a contextual representation rather than as a geometrically registered object-level feature. As discussed in Section~\ref{sec:crossmodal_observation}, visual and forward-looking sonar observations are formed in different imaging geometries. UMOD's optical and sonar sensors provide temporally synchronized measurements, but their image coordinates are not directly interchangeable; visual observations are represented in an image plane, whereas the forward-looking sonar encodes acoustic returns in range-azimuth coordinates \cite{wu2026multimodal}. The original UMOD study similarly identifies cross-modal spatial misalignment as a central difficulty for visual-sonar feature fusion. We therefore do not transfer a visual bounding box directly into the sonar image.

Instead, the complete synchronized sonar frame is encoded to obtain an acoustic context vector:
\begin{equation}
    \mathbf{z}_s
    =
    f_{\mathrm{sonar}}\!\left(\mathbf{x}_s\right).
\end{equation}
The sonar encoder is an ImageNet-pretrained ResNet-18 used as a frozen feature extractor. Its final classification layer is removed, and the pooled representation preceding the classifier is retained, yielding
\begin{equation}
    \mathbf{z}_s
    \in
    \mathbb{R}^{512}.
\end{equation}

The sonar encoder is held fixed in the same manner as the visual encoder. Furthermore, because the degradation benchmark modifies only the optical modality,
\begin{equation}
    \mathbf{z}_s^{(0)}
    =
    \mathbf{z}_s^{(1)}
    =
    \cdots
    =
    \mathbf{z}_s^{(4)}
\end{equation}
for all degradation variants derived from the same synchronized observation. Thus, the acoustic representation serves as a controlled source of complementary information whose input quality is not altered by the visual degradation procedure.

This design should be distinguished from object-level visual-sonar alignment. The sonar feature may contain information from the target, surrounding scene structure, background acoustic returns, or other contextual characteristics of the synchronized observation. Accordingly, the present study evaluates whether \emph{paired acoustic context} can compensate for deteriorating visual evidence. Explicit object correspondence and spatially aligned visual-sonar feature extraction are left to future end-to-end extensions.

\subsection{Fixed Cross-Modal Fusion}
\label{sec:fixed_fusion}

We first establish a fixed multimodal baseline to determine whether the availability of sonar information alone is sufficient to improve robustness. A straightforward feature-level fusion strategy is constructed by concatenating the frozen visual and sonar embeddings:
\begin{equation}
    \mathbf{z}_{\mathrm{cat}}^{(s)}
    =
    \left[
        \mathbf{z}_v^{(s)};
        \mathbf{z}_s
    \right],
\end{equation}
where $[\cdot;\cdot]$ denotes vector concatenation. Given the $384$-dimensional visual representation and $512$-dimensional sonar representation,
\begin{equation}
    \mathbf{z}_{\mathrm{cat}}^{(s)}
    \in
    \mathbb{R}^{896}.
\end{equation}

A downstream classifier $h_{\mathrm{fix}}(\cdot)$ maps the concatenated representation to the retained target classes:
\begin{equation}
    \hat{y}_{\mathrm{fix}}^{(s)}
    =
    h_{\mathrm{fix}}
    \left(
        \mathbf{z}_{\mathrm{cat}}^{(s)}
    \right).
\end{equation}

The fixed-fusion baseline provides no explicit mechanism for modifying the contribution of either modality as visual conditions change. Consequently, features extracted from a severely degraded optical observation are presented to the classifier in the same manner as features extracted under clean conditions. This baseline is particularly relevant because previous experiments on UMOD demonstrate that direct addition or concatenation of heterogeneous visual and sonar features can introduce inter-modal interference rather than guarantee an improvement over unimodal perception \cite{wu2026multimodal}. The fixed model therefore tests whether complementary acoustic information is intrinsically sufficient for robustness or whether explicit adaptation of modality contributions is required.

\subsection{Adaptive Gated Fusion}
\label{sec:gated_fusion}

To permit the model to vary its reliance on each sensing modality, we introduce a lightweight adaptive gated fusion mechanism. Because the DINOv2 and sonar encoders produce representations with different dimensions and feature statistics, the two embeddings are first mapped into a common latent space. Let $P_v(\cdot)$ and $P_s(\cdot)$ denote learnable modality-specific projection functions. The projected representations are
\begin{equation}
    \bar{\mathbf{z}}_v^{(s)}
    =
    P_v\!\left(\mathbf{z}_v^{(s)}\right),
    \qquad
    \bar{\mathbf{z}}_s
    =
    P_s\!\left(\mathbf{z}_s\right),
\end{equation}
with
\begin{equation}
    \bar{\mathbf{z}}_v^{(s)},
    \bar{\mathbf{z}}_s
    \in
    \mathbb{R}^{d},
\end{equation}
where $d$ denotes the shared fusion dimension.

The projected representations are concatenated and passed to a gating network $G(\cdot)$:
\begin{equation}
    \mathbf{a}^{(s)}
    =
    G
    \left(
        \left[
            \bar{\mathbf{z}}_v^{(s)};
            \bar{\mathbf{z}}_s
        \right]
    \right),
\end{equation}
where
\begin{equation}
    \mathbf{a}^{(s)}
    =
    \left[
        a_v^{(s)},a_s^{(s)}
    \right].
\end{equation}
A softmax operation converts these logits into normalized modality weights:
\begin{equation}
    \left[
        g_v^{(s)},g_s^{(s)}
    \right]
    =
    \operatorname{softmax}
    \left(
        \mathbf{a}^{(s)}
    \right),
\end{equation}
such that
\begin{equation}
    g_v^{(s)} \geq 0,
    \qquad
    g_s^{(s)} \geq 0,
    \qquad
    g_v^{(s)}+g_s^{(s)}=1.
\end{equation}

The fused representation is then constructed as a weighted combination of the two projected embeddings:
\begin{equation}
    \mathbf{z}_{g}^{(s)}
    =
    g_v^{(s)}
    \bar{\mathbf{z}}_v^{(s)}
    +
    g_s^{(s)}
    \bar{\mathbf{z}}_s.
\end{equation}
Finally, the class prediction is obtained from
\begin{equation}
    \hat{y}_{g}^{(s)}
    =
    h_g
    \left(
        \mathbf{z}_{g}^{(s)}
    \right),
\end{equation}
where $h_g(\cdot)$ is the learned classification head.

Unlike fixed concatenation, the gated formulation makes the effective contribution of each modality sample dependent. In particular, changes in the visual representation caused by degradation can alter the gating function and thereby change the relative contribution assigned to the acoustic branch. The model is not explicitly supplied with the degradation level $s$ as an input; instead, the gate must infer an appropriate weighting from the representations themselves.

The learned coefficients should not be interpreted as calibrated probabilities that a physical sensor is reliable. Rather, $g_v^{(s)}$ and $g_s^{(s)}$ quantify the relative contribution assigned by the learned fusion mechanism to the two projected representations. Their variation across degradation levels nevertheless provides a useful diagnostic for determining whether the model changes its use of acoustic information as the visual observation becomes less informative.

\subsection{Degradation-Aware Fusion Training}
\label{sec:degradation_aware_training}

Adaptive gating alone does not ensure robustness to changes in modality reliability. If the gating mechanism is optimized exclusively on clean visual observations, the training distribution provides little incentive to learn how modality contributions should change when the optical representation becomes unreliable. We therefore distinguish between \emph{clean-trained adaptive fusion} and \emph{degradation-aware adaptive fusion}.

For clean-trained fusion, the gating network, projection layers, and classifier are optimized using only the original $D_0$ visual observations:
\begin{equation}
    \mathcal{T}_{\mathrm{clean}}
    =
    \left\{
        \left(
            \mathbf{z}_{v,i}^{(0)},
            \mathbf{z}_{s,i},
            y_i
        \right)
    \right\}_{i=1}^{N_{\mathrm{tr}}},
\end{equation}
where $N_{\mathrm{tr}}$ denotes the number of object-level training examples. This model has access to both modalities during training but does not explicitly observe systematic reductions in visual quality.

For degradation-aware training, every training object is evaluated across the complete visual degradation sequence. The training set becomes
\begin{equation}
    \mathcal{T}_{\mathrm{DA}}
    =
    \bigcup_{s=0}^{4}
    \left\{
        \left(
            \mathbf{z}_{v,i}^{(s)},
            \mathbf{z}_{s,i},
            y_i
        \right)
    \right\}_{i=1}^{N_{\mathrm{tr}}}.
\end{equation}
Importantly, all five visual variants associated with a training object are paired with the same sonar representation and class label. The procedure therefore varies the reliability of the optical evidence while preserving sample identity and acoustic context. No degraded version of a validation or test observation is introduced into the training set, consistent with the grouped partitioning procedure described in Section~\ref{sec:dataset_setup}.

The trainable parameters of the projection layers, gating network, and classification head are optimized using the multiclass cross-entropy objective
\begin{equation}
    \mathcal{L}_{\mathrm{cls}}
    =
    -
    \frac{1}{N}
    \sum_{i=1}^{N}
    \sum_{c=1}^{C}
    \mathbb{I}(y_i=c)
    \log
    p_{i,c},
\end{equation}
where $p_{i,c}$ denotes the predicted probability that observation $i$ belongs to class $c$. The parameters of the DINOv2 and sonar encoders remain frozen during this optimization. Thus, degradation-aware learning acts specifically on the mapping and fusion components rather than modifying the underlying pretrained representations.

The distinction between the two gating regimes constitutes an important ablation. Both models have identical access to visual and sonar information and use the same adaptive fusion formulation; they differ only in whether the fusion mechanism is exposed to changing visual reliability during optimization. Consequently, a robustness advantage for degradation-aware gating can be attributed to training across reliability conditions rather than to the existence of a gating architecture alone.

More generally, degradation-aware training can be viewed as learning a conditional fusion function
\begin{equation}
    \Phi_{\mathrm{DA}}
    :
    \left(
        \mathbf{z}_v^{(s)},
        \mathbf{z}_s
    \right)
    \mapsto
    \mathbf{z}_{g}^{(s)}
\end{equation}
whose behavior is allowed to vary with the evidence contained in each modality. The desired behavior is not to force monotonically increasing sonar weight as a hard constraint, but to allow such a shift to emerge when acoustic context becomes more useful relative to the degraded visual representation. The resulting modality weights are therefore analyzed empirically across $D_0$-$D_4$ in the experiments that follow.

Overall, the proposed methodology separates three questions that would otherwise be confounded in a fully end-to-end multimodal architecture: whether a frozen foundation-model representation retains discriminative information under underwater degradation, whether synchronized sonar context provides complementary information, and whether exposure to changing visual reliability is necessary for an adaptive fusion mechanism to exploit that complementarity. This controlled design provides the basis for the robustness comparisons presented in the following section.

\section{Experiments and Results}
\label{sec:experiments_results}

This section evaluates the robustness of visual and cross-modal underwater perception under the controlled degradation protocol introduced in Section~\ref{sec:degradation_model}. The experiments are organized to progressively address three questions: (i) how rapidly conventional detection and frozen foundation-model representations deteriorate as visual quality decreases, (ii) whether the availability of synchronized sonar context is by itself sufficient to improve robustness, and (iii) whether explicitly exposing an adaptive fusion mechanism to changes in visual reliability enables more effective cross-modal compensation. Unless otherwise stated, recognition results are reported using balanced accuracy on the fixed test partition, while visual detection is evaluated using mAP. As discussed in Section~\ref{sec:evaluation_metrics}, the absolute scores of the detection and recognition tasks are not directly comparable; comparisons across these tasks therefore emphasize their relative degradation trends.

\subsection{Visual Perception under Progressive Degradation}
\label{sec:visual_degradation_results}

We first characterize the behavior of visual-only perception before introducing acoustic information. Two complementary baselines are considered: YOLO11n operating as an end-to-end object detector and frozen DINOv2-small representations evaluated through object-crop classification. The former measures the combined sensitivity of localization and classification to visual corruption, whereas the latter isolates the robustness of pretrained visual representations when object localization is provided by the ground-truth annotation.

Under the clean $D_0$ condition, YOLO11n achieves an $\mathrm{mAP}@0.5$ of $0.7020$ and an $\mathrm{mAP}@0.5{:}0.95$ of $0.5063$. Performance decreases progressively as degradation becomes stronger. As shown in Table~\ref{tab:yolo_degradation}, $\mathrm{mAP}@0.5$ decreases from $0.7020$ at $D_0$ to $0.6215$, $0.5676$, and $0.3359$ at $D_1$, $D_2$, and $D_3$, respectively. Under the extreme $D_4$ condition, the detector retains only $0.0281$ $\mathrm{mAP}@0.5$, corresponding to approximately $4.0\%$ of its $D_0$ performance. Recall similarly decreases from $0.5127$ at $D_0$ to $0.0215$ at $D_4$, indicating that the principal failure mode under extreme degradation is the loss of detectable visual evidence rather than merely reduced localization precision.

\begin{table}[!htb]
\centering
\caption{YOLO11n visual detection performance under progressive combined degradation.}
\label{tab:yolo_degradation}
\begin{tabular}{c c c c c}
\hline
Condition & mAP@0.5 & mAP@0.5:0.95 & Precision & Recall \\
\hline
$D_0$ & 0.7020 & 0.5063 & 0.8299 & 0.5127 \\
$D_1$ & 0.6215 & 0.4538 & 0.6860 & 0.5606 \\
$D_2$ & 0.5676 & 0.4319 & 0.6777 & 0.5001 \\
$D_3$ & 0.3359 & 0.2366 & 0.8492 & 0.2952 \\
$D_4$ & 0.0281 & 0.0219 & 0.8000 & 0.0215 \\
\hline
\end{tabular}
\end{table}

The frozen DINOv2 representation exhibits substantially greater relative stability. Its balanced accuracy is $0.6210$ at $D_0$, increases slightly to $0.6495$ at $D_1$, and remains at $0.6095$ under $D_2$. More substantial degradation appears at $D_3$ and $D_4$, where balanced accuracy decreases to $0.4952$ and $0.4610$, respectively. Thus, even at the most severe combined degradation level, the foundation-model representation retains approximately $74.2\%$ of its clean-condition balanced accuracy. This result is consistent with the motivation for using large-scale self-supervised representations as a comparatively stable visual feature space under distribution shift \cite{oquab2023dinov2,hendrycks2021pretraining,paul2022vision}. Nevertheless, the decline from $0.6210$ to $0.4610$ also demonstrates that representation robustness does not eliminate the effects of severe information loss.

\begin{figure}[!htb]
    \centering
    \includegraphics[width=\columnwidth]{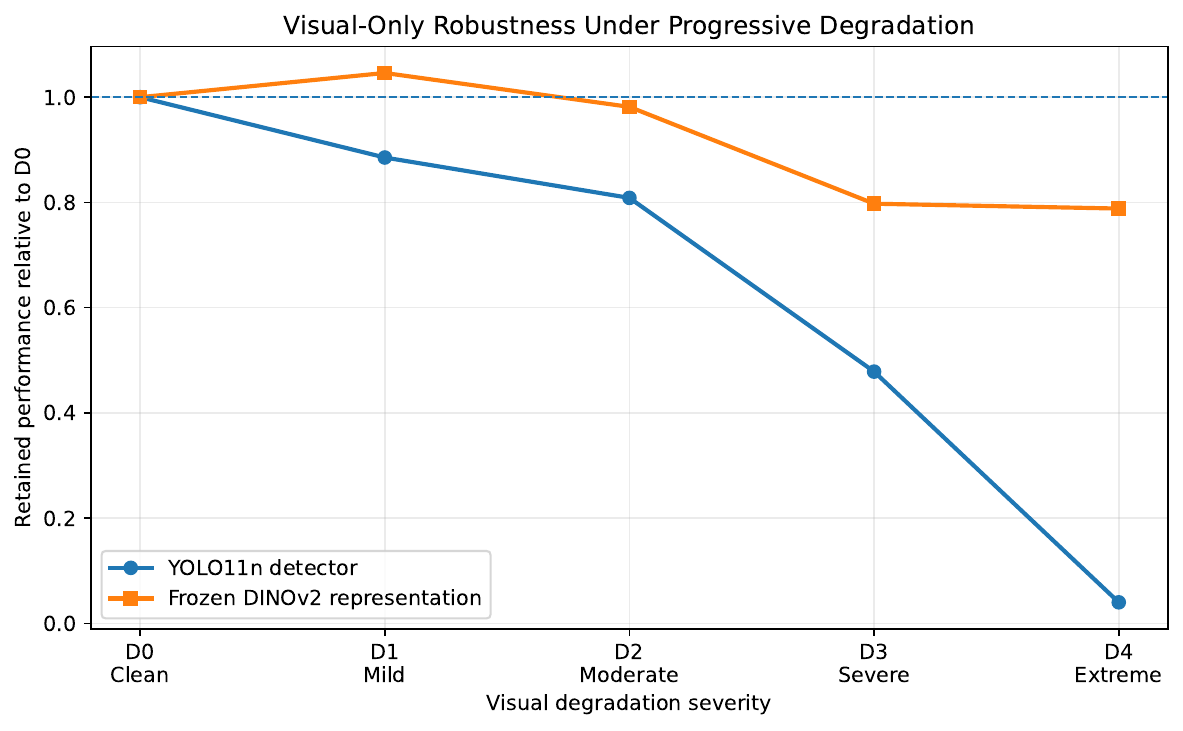}
    \caption{Relative robustness of visual-only perception under progressive degradation. Each curve is normalized to its own $D_0$ performance because YOLO11n and DINOv2 are evaluated on different tasks and metrics. The detector exhibits a sharp degradation at $D_3$-$D_4$, whereas the frozen DINOv2 representation retains substantially more of its clean-condition performance.}
    \label{fig:visual_robustness}
\end{figure}

Figure~\ref{fig:visual_robustness} summarizes the normalized degradation behavior of the two visual approaches. YOLO11n obtains an MRR of $0.5531$, whereas the frozen DINOv2 representation obtains an MRR of $0.8919$ under the final evaluation protocol. These values should be interpreted as performance retention within each task rather than as a direct comparison between detection accuracy and object-level classification accuracy. In particular, YOLO11n must localize degraded objects in addition to recognizing them, while the DINOv2 experiment uses ground-truth object crops. The result therefore supports the narrower conclusion that the frozen foundation-model representation is comparatively invariant to the tested perturbations once object localization is supplied.

\subsection{Effect of Sonar Context and Fixed Fusion}
\label{sec:fixed_fusion_results}

We next examine whether the presence of synchronized sonar information automatically improves robustness. Three object-level recognition configurations are compared: the visual DINOv2 baseline, sonar context alone, and fixed visual-sonar feature concatenation. Because only the optical observation is modified across $D_0$-$D_4$, the sonar-only classifier provides a constant reference point with balanced accuracy $0.3581$ across all five conditions.

Table~\ref{tab:crossmodal_results} reports the complete cross-modal comparison. Under clean conditions, fixed fusion achieves a balanced accuracy of $0.6457$, modestly exceeding the visual-only score of $0.6210$. At $D_1$, the fixed-fusion score increases to $0.7143$, compared with $0.6495$ for the visual representation. This advantage, however, is not preserved as visual reliability becomes severely degraded. At $D_3$, fixed fusion reaches $0.5162$, only slightly above the visual-only result of $0.4952$. At $D_4$, its balanced accuracy decreases to $0.4343$, below the visual-only score of $0.4610$.

\begin{table*}[!htb]
\centering
\caption{Balanced accuracy of object-level visual, acoustic, and cross-modal recognition models under progressive combined degradation.}
\label{tab:crossmodal_results}
\begin{tabular}{c c c c c c}
\hline
Condition & Visual FM & Sonar Context & Fixed Fusion & Clean-Trained Gate & Degradation-Aware Gate \\
\hline
$D_0$ & 0.6210 & 0.3581 & 0.6457 & 0.6210 & 0.6343 \\
$D_1$ & 0.6495 & 0.3581 & 0.7143 & 0.6610 & 0.6743 \\
$D_2$ & 0.6095 & 0.3581 & 0.6152 & 0.7010 & 0.6743 \\
$D_3$ & 0.4952 & 0.3581 & 0.5162 & 0.5010 & \textbf{0.6438} \\
$D_4$ & 0.4610 & 0.3581 & 0.4343 & 0.3752 & \textbf{0.6152} \\
\hline
\end{tabular}
\end{table*}

The behavior of fixed concatenation is important because it demonstrates that complementary sensing does not imply complementary representations at every operating condition. The sonar feature is available at $D_4$, yet direct concatenation fails to exploit it sufficiently to offset the deteriorating visual representation. This observation is consistent with previous findings on UMOD, where simple addition and concatenation of heterogeneous visual and sonar features reduced detection performance relative to a unimodal visual baseline, motivating more structured cross-modal interaction \cite{wu2026multimodal}. In the present setting, the same general issue appears under a different experimental condition: as the reliability of the visual modality changes, a fusion rule learned without an explicit mechanism for reliability adaptation can become less effective than the visual representation alone.

The result also clarifies the role of the sonar-only baseline. Although its balanced accuracy of $0.3581$ is below the clean visual score, its performance is invariant to the imposed optical degradation. Consequently, the acoustic representation becomes relatively more informative as the visual modality deteriorates. The central problem is therefore not whether sonar alone is more discriminative than vision, but whether the fusion mechanism can recognize when the relative utility of the two modalities has changed.

\subsection{Degradation-Aware Adaptive Fusion}
\label{sec:degradation_aware_results}

To determine whether adaptive weighting alone is sufficient, we compare fixed fusion with two instances of the gated model described in Section~\ref{sec:gated_fusion}. The \emph{clean-trained gate} is optimized using only $D_0$ visual observations, whereas the \emph{degradation-aware gate} is optimized using visual representations spanning the complete $D_0$-$D_4$ range. The two gated models otherwise employ the same fusion formulation, allowing the effect of degradation-aware training to be isolated.

The clean-trained gate performs competitively under mild and moderate conditions. Its balanced accuracy increases from $0.6210$ at $D_0$ to $0.6610$ at $D_1$ and $0.7010$ at $D_2$. However, this behavior does not extend to severe reliability shifts. Performance falls to $0.5010$ at $D_3$ and $0.3752$ at $D_4$. The latter is lower than both the visual-only baseline ($0.4610$) and fixed fusion ($0.4343$). Therefore, the existence of a learnable gating architecture does not by itself guarantee robustness to a modality condition absent from its training distribution.

In contrast, degradation-aware fusion remains substantially more stable as visual reliability decreases. Its balanced accuracy is $0.6343$ under clean conditions and $0.6743$ under both $D_1$ and $D_2$. More importantly, it maintains $0.6438$ at $D_3$ and $0.6152$ at $D_4$. Relative to the visual-only foundation-model baseline, the $D_4$ improvement is
\begin{equation}
    0.6152 - 0.4610 = 0.1543,
\end{equation}
corresponding to a relative improvement of
\begin{equation}
    \frac{0.6152-0.4610}{0.4610}\times100
    \approx 33.5\%.
\end{equation}

\begin{figure}[!htb]
    \centering
    \includegraphics[width=0.5\textwidth]{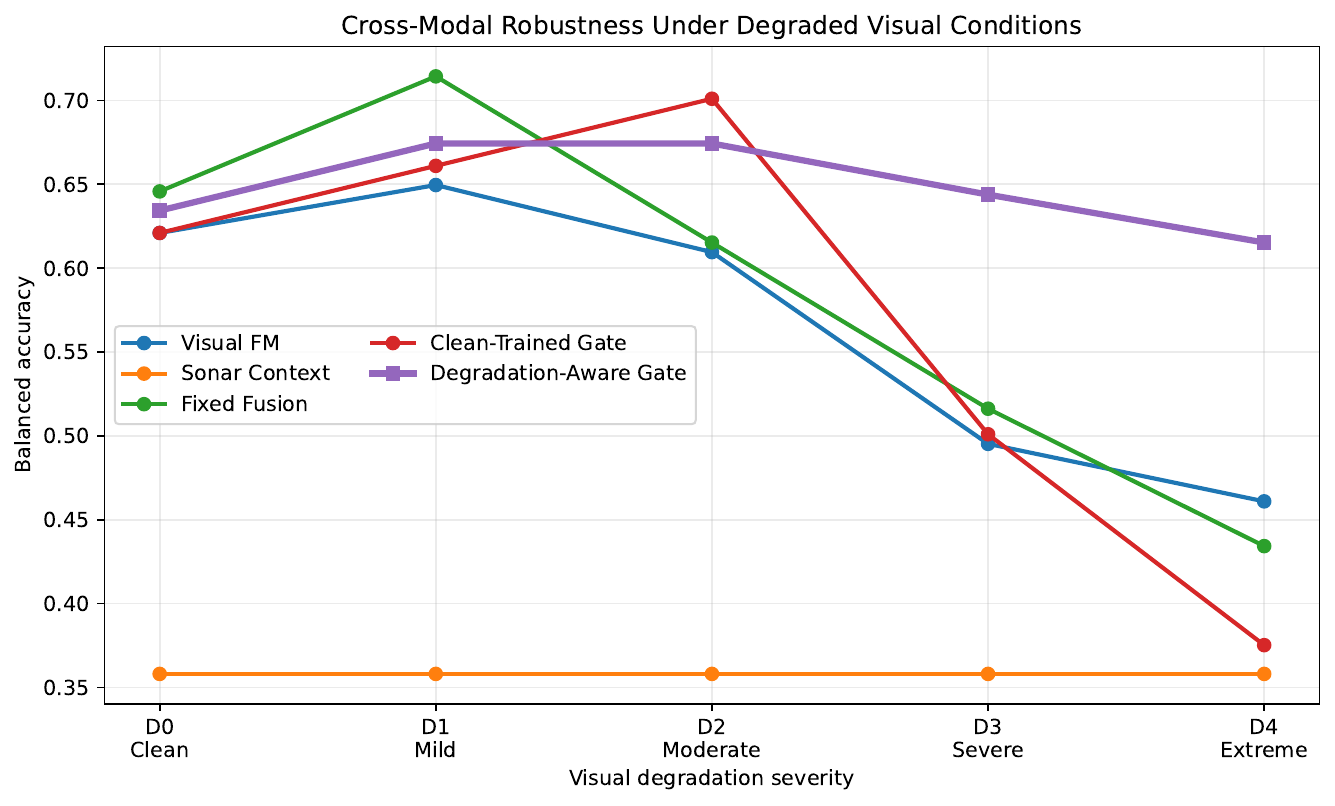}
    \caption{Cross-modal recognition performance under progressive visual degradation. Fixed fusion and clean-trained adaptive gating provide inconsistent robustness as visual reliability decreases. In contrast, degradation-aware gating maintains substantially higher balanced accuracy under severe ($D_3$) and extreme ($D_4$) degradation while preserving competitive clean-condition performance.}
    \label{fig:crossmodal_robustness}
\end{figure}

At $D_3$, the improvement is similarly substantial: degradation-aware fusion reaches $0.6438$ compared with $0.4952$ for the visual-only model. Notably, the degradation-aware model also exceeds both fixed fusion ($0.5162$) and the clean-trained gate ($0.5010$) at this severity. Figure~\ref{fig:crossmodal_robustness} illustrates this divergence as the degradation becomes severe.

The robustness summary in Table~\ref{tab:robustness_summary} further emphasizes this behavior. Fixed fusion retains approximately $67.3\%$ of its clean balanced accuracy at $D_4$, while the clean-trained gate retains approximately $60.4\%$. The degradation-aware gate retains approximately $97.0\%$ of its $D_0$ score and obtains an MRR of $1.0278$. The latter should not be interpreted as evidence that degradation improves the underlying observations. Rather, as discussed in Section~\ref{sec:robustness_objective}, MRR is normalized to the model's own $D_0$ performance, and small non-monotonic variations can occur on the limited test set. The relevant observation is that degradation-aware fusion exhibits substantially less performance loss across the tested reliability shift. The distinction becomes particularly evident under severe degradation. As shown in Fig.~\ref{fig:crossmodal_robustness}, the performance trajectories of the nominally trained models begin to diverge sharply from degradation-aware fusion at $D_3$. At $D_4$, degradation-aware fusion maintains a balanced accuracy of $0.6152$, compared with $0.4610$ for the Visual FM, $0.4343$ for fixed fusion, and $0.3752$ for the clean-trained gate.

\begin{table}[!htb]
\centering
\caption{Robustness summary for the object-level recognition models.}
\label{tab:robustness_summary}
\begin{tabular}{l c c c}
\hline
Model & $D_4$ BAcc & $D_4$ Retained & MRR \\
\hline
Visual FM & 0.4610 & 0.7423 & 0.8919 \\
Sonar Context & 0.3581 & 1.0000 & 1.0000 \\
Fixed Fusion & 0.4343 & 0.6726 & 0.8827 \\
Clean-Trained Gate & 0.3752 & 0.6043 & 0.9011 \\
Degradation-Aware Gate & \textbf{0.6152} & \textbf{0.9700} & \textbf{1.0278} \\
\hline
\end{tabular}
\end{table}

Taken together, these results support the principal hypothesis of this study: adaptive architecture and multimodal availability are insufficient when the fusion model has not experienced changes in modality reliability. Robustness emerges most clearly when the fusion mechanism is trained across the reliability conditions on which it is expected to operate. This observation is closely related to broader findings in robust multimodal learning, where explicit exposure to missing or corrupted modalities improves the ability of multimodal models to avoid dependence on a consistently high-quality input \cite{neverova2016moddrop,ma2022smil}.

\subsection{Adaptation of Modality Reliance}
\label{sec:modality_weight_results}

\begin{figure}[!htb]
    \centering
    \includegraphics[width=\columnwidth]{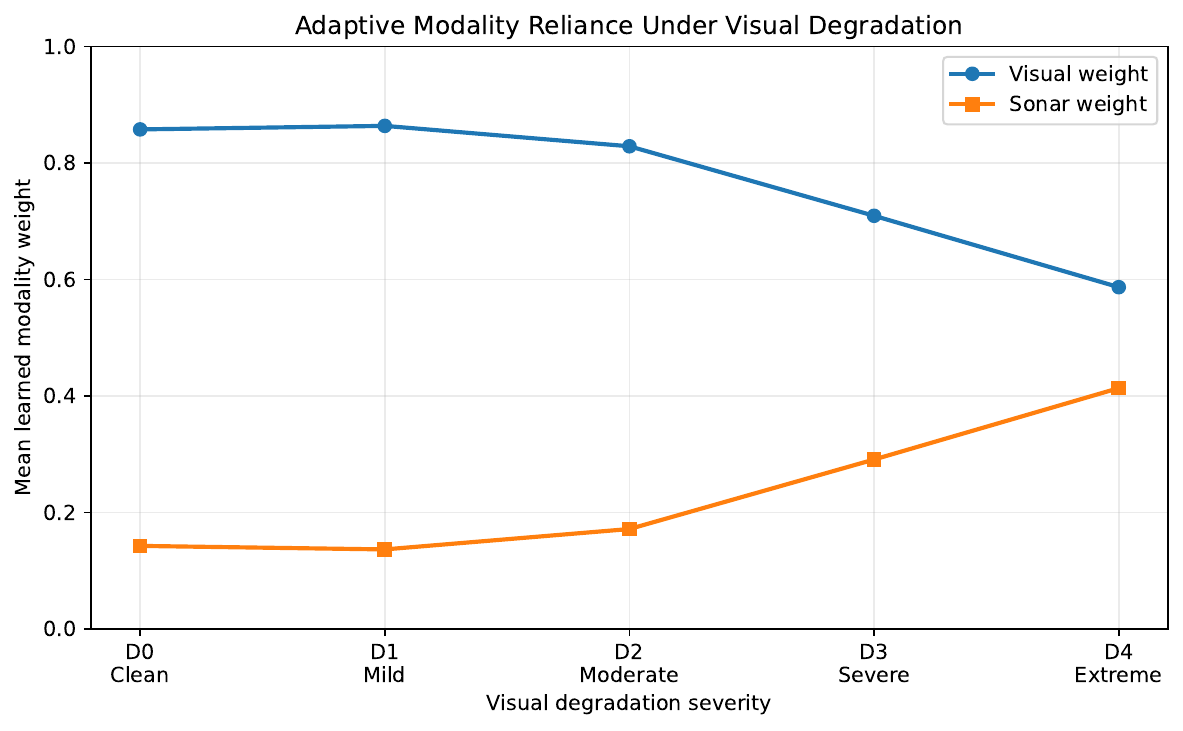}
    \caption{Mean modality weights learned by degradation-aware gated fusion. The model remains predominantly visual under clean and mildly degraded conditions but progressively increases the contribution of sonar as optical degradation becomes severe. The coefficients represent learned fusion contributions and should not be interpreted as calibrated sensor-reliability probabilities.}
    \label{fig:modality_weights}
\end{figure}

The previous experiment establishes that degradation-aware fusion improves predictive robustness. We next examine whether this improvement is accompanied by the expected change in modality use. For each degradation level, we average the learned visual and sonar gating coefficients over the test objects. These coefficients represent the relative contributions assigned by the learned fusion mechanism and, as emphasized in Section~\ref{sec:gated_fusion}, should not be interpreted as calibrated sensor-reliability probabilities.

Table~\ref{tab:modality_weights} shows a clear shift toward acoustic information as the visual degradation becomes severe. Under $D_0$, the average visual and sonar weights are $0.8576$ and $0.1424$, respectively. The gate remains strongly vision dominated under mild degradation, assigning a sonar weight of $0.1363$ at $D_1$. At $D_2$, the sonar contribution increases modestly to $0.1713$. A much larger redistribution occurs at $D_3$ and $D_4$: the average sonar weight rises to $0.2907$ and $0.4133$, respectively, while the corresponding visual weight decreases to $0.7093$ and $0.5867$.

\begin{table}[!htb]
\centering
\caption{Average modality weights learned by the degradation-aware gate.}
\label{tab:modality_weights}
\begin{tabular}{c c c}
\hline
Condition & Visual Weight & Sonar Weight \\
\hline
$D_0$ & 0.8576 & 0.1424 \\
$D_1$ & 0.8637 & 0.1363 \\
$D_2$ & 0.8287 & 0.1713 \\
$D_3$ & 0.7093 & 0.2907 \\
$D_4$ & 0.5867 & 0.4133 \\
\hline
\end{tabular}
\end{table}

From $D_0$ to $D_4$, the mean sonar contribution therefore increases by approximately $0.271$, or $27.1$ percentage points. Importantly, monotonic behavior was not imposed as a training constraint, nor was the degradation severity supplied explicitly to the gate. The small decrease in sonar weight between $D_0$ and $D_1$ further confirms that the learned coefficients are not simply a deterministic function of the predefined severity index. Instead, the pronounced increase at $D_3$ and $D_4$ emerges from the representations presented to the fusion network.

Figure~\ref{fig:modality_weights} visualizes this transition. The result provides a mechanistic explanation for the robustness improvement observed in Section~\ref{sec:degradation_aware_results}: when the visual representation becomes substantially less discriminative, the degradation-aware model does not continue to use the two modalities in approximately the same manner as under clean conditions. Instead, it increases the contribution of the unchanged acoustic context while still retaining a majority visual contribution. Thus, the model behaves as a complementary fusion system rather than simply replacing vision with sonar.

\subsection{Degradation-Type Ablation}
\label{sec:degradation_type_results}

The combined $D_0$-$D_4$ benchmark changes several optical properties simultaneously. To determine which perturbations are primarily responsible for the benefit of acoustic assistance, we separately evaluate illumination reduction, wavelength-dependent color attenuation, turbidity, and blur at moderate ($D_2$) and extreme ($D_4$) severity. Table~\ref{tab:degradation_ablation} compares the visual foundation-model baseline with degradation-aware fusion.

\begin{table*}[!htb]
\centering
\caption{Degradation-type ablation. Relative improvement is computed with respect to the visual foundation-model balanced accuracy at the same degradation type and severity.}
\label{tab:degradation_ablation}
\begin{tabular}{c c c c c c}
\hline
Degradation & Severity & Visual FM & DA Fusion & Absolute Gain & Relative Gain (\%) \\
\hline
Brightness & $D_2$ & 0.6495 & 0.6743 & 0.0248 & 3.81 \\
Brightness & $D_4$ & 0.6095 & 0.6457 & 0.0362 & 5.94 \\
Color & $D_2$ & 0.6210 & 0.6743 & 0.0533 & 8.59 \\
Color & $D_4$ & 0.7295 & 0.7295 & 0.0000 & 0.00 \\
Turbidity & $D_2$ & 0.5924 & 0.6457 & 0.0533 & 9.00 \\
Turbidity & $D_4$ & 0.4667 & 0.5638 & 0.0971 & \textbf{20.82} \\
Blur & $D_2$ & 0.6629 & 0.7029 & 0.0400 & 6.03 \\
Blur & $D_4$ & 0.6190 & 0.7143 & 0.0952 & \textbf{15.38} \\
\hline
\end{tabular}
\end{table*}

The largest cross-modal benefit occurs under severe turbidity. At $D_4$, the visual-only balanced accuracy is $0.4667$, whereas degradation-aware fusion reaches $0.5638$, yielding an absolute gain of $0.0971$ and a relative improvement of $20.8\%$. Severe blur exhibits the second-largest benefit: balanced accuracy increases from $0.6190$ to $0.7143$, corresponding to a $15.4\%$ relative improvement. These perturbations directly suppress spatial structure, object boundaries, contrast, and fine-scale visual evidence, which are important cues for object recognition. Because acoustic sensing is not affected by optical scattering or image-plane blur, the synchronized sonar representation can provide complementary information when such visual structure becomes unreliable.

Brightness degradation produces a smaller but consistent benefit. At $D_4$, fusion improves balanced accuracy from $0.6095$ to $0.6457$, a relative increase of approximately $5.9\%$. Color attenuation behaves differently. Although fusion improves the $D_2$ result from $0.6210$ to $0.6743$, both models obtain $0.7295$ under the isolated $D_4$ color perturbation. Thus, no measurable fusion gain is observed in that condition. The relatively strong visual-only performance suggests that the frozen DINOv2 representation remains discriminative when the perturbation primarily changes channel statistics without equivalently destroying spatial structure.

These degradation-specific results refine the interpretation of the combined benchmark. Sonar assistance is not uniformly advantageous for every type of optical shift. Instead, its largest contribution appears when the perturbation removes or obscures structural visual evidence, particularly under turbidity and blur. This distinction is consistent with the complementary sensing characteristics of optical and acoustic imagery: optical observations provide rich appearance information but are sensitive to visibility, whereas sonar preserves acoustic structural information under conditions in which optical propagation is compromised \cite{wu2026multimodal}.

\subsection{Qualitative Failure Recovery}
\label{sec:qualitative_results}

Finally, we examine individual $D_4$ test objects to determine whether the aggregate improvement of degradation-aware fusion corresponds to concrete recovery of visual-only failures. Across the extreme combined-degradation test set, eight objects that are incorrectly classified by the visual foundation-model baseline are correctly classified after degradation-aware visual-sonar fusion. 

These cases provide qualitative evidence for the behavior observed quantitatively in the preceding experiments. Under extreme degradation, the visual crop may contain substantially reduced contrast, attenuated color information, scattering-induced veiling, and blurred object structure. The paired sonar observation remains unchanged, and the learned gate assigns substantially greater acoustic contribution under these conditions. In the illustrated recovery cases, this additional acoustic context is sufficient to alter the final decision toward the correct class.

The qualitative examples should not be interpreted independently as statistical evidence of superiority, particularly given the limited size of the present test set. Rather, they illustrate how the quantitative $D_4$ improvement manifests at the sample level. Together with the $33.5\%$ relative balanced-accuracy improvement and the increase in mean sonar weight from $14.2\%$ to $41.3\%$, the recovered examples support the interpretation that degradation-aware training enables the fusion mechanism to make productive use of acoustic context when visual evidence becomes severely compromised.

\subsection{Summary of Experimental Findings}
\label{sec:experimental_summary}

The experiments reveal three principal findings. First, progressive underwater degradation strongly affects conventional visual detection, while frozen DINOv2 representations exhibit substantially greater relative stability. However, the foundation-model representation still loses discriminative performance under severe and extreme degradation, indicating that pretrained visual invariance alone is insufficient when optical information is substantially removed.

Second, simply adding sonar information does not guarantee robustness. Fixed concatenation achieves useful gains under some mild conditions but falls below the visual-only baseline at $D_4$. Similarly, an adaptive gate trained only on clean observations deteriorates sharply under extreme degradation. These results reinforce the distinction between \emph{multimodal availability} and \emph{multimodal robustness}: a complementary modality can only be useful if the fusion mechanism learns how to exploit it when the reliability of the primary modality changes.

Third, degradation-aware training substantially changes this behavior. The proposed gate maintains a balanced accuracy of $0.6152$ at $D_4$, compared with $0.4610$ for visual-only recognition, while increasing the average sonar contribution from $14.2\%$ at $D_0$ to $41.3\%$ at $D_4$. The benefit is degradation dependent, with the largest isolated gains observed under severe turbidity ($20.8\%$) and blur ($15.4\%$). Collectively, these findings indicate that robust visual-sonar perception is determined not only by the complementarity of the sensing modalities, but also by whether the fusion mechanism is explicitly trained to operate across changes in their relative reliability.

\section{Discussion}
\label{sec:discussion}

The experimental results provide several insights into robust underwater perception under changing optical reliability. The central finding is that robustness cannot be attributed solely to either a strong visual representation or the availability of a complementary sensing modality. Instead, the results indicate that reliable cross-modal perception depends on the interaction between representation robustness, modality complementarity, and the conditions under which the fusion mechanism is trained. In particular, the degradation-aware gated model substantially outperforms both visual-only perception and alternative fusion strategies under severe degradation while maintaining competitive performance under clean and moderately degraded conditions. This section discusses the implications of these findings, their relationship to prior work, and the principal limitations of the present study.

\subsection{Visual Foundation Models Improve Robustness but Do Not Eliminate Information Loss}
\label{sec:discussion_visual_fm}

A first observation is the markedly different degradation behavior of conventional end-to-end detection and frozen foundation-model representations. The YOLO11n detector experiences a substantial decrease in performance as the combined optical degradation progresses, with $\mathrm{mAP}@0.5$ decreasing from $0.7020$ under $D_0$ to $0.0281$ under $D_4$. By contrast, the frozen DINOv2 representation retains substantially more of its clean-condition recognition performance, achieving a balanced accuracy of $0.4610$ under extreme combined degradation compared with $0.6210$ under clean conditions.

This result is consistent with previous evidence that large-scale self-supervised and transformer-based visual representations can exhibit improved stability under distribution shift relative to conventional task-specific representations \cite{oquab2023dinov2,hendrycks2021pretraining,paul2022vision}. The DINOv2 encoder was not fine-tuned on the degradation benchmark and therefore did not explicitly learn the perturbations considered in this work. Its relative stability consequently suggests that large-scale pretraining provides useful invariances that transfer, at least partially, to underwater imagery outside the original pretraining distribution.

Nevertheless, the extreme-degradation results also identify an important limitation of foundation-model robustness. A pretrained representation may suppress sensitivity to nuisance variations, but it cannot reconstruct visual evidence that has been physically attenuated, scattered, or blurred beyond recoverability. Underwater image formation is fundamentally affected by wavelength-dependent attenuation, backscatter, and distance-dependent transmission, and these effects can remove information rather than merely alter its statistical appearance \cite{akkaynak2018revised,akkaynak2019seathru}. Accordingly, the observed reduction in DINOv2 balanced accuracy under $D_3$ and $D_4$ is not unexpected. The result suggests that foundation models should be regarded as a strong component of robust underwater perception rather than as a substitute for complementary sensing.

This distinction is particularly relevant for marine robotics. Foundation models have been proposed as broadly transferable components capable of supporting perception and reasoning across multiple downstream tasks \cite{bommasani2021opportunities}. However, the present results indicate that their deployment in underwater environments must account for sensor-specific physical limitations. Robust representation learning can reduce sensitivity to moderate domain shifts, but multimodal sensing remains important when one modality undergoes severe information loss.

\subsection{Complementary Sensors Do Not Guarantee Robust Fusion}
\label{sec:discussion_complementarity}

The original UMOD study demonstrated that simply adding or concatenating heterogeneous visual and sonar features can introduce inter-modal interference; our Fig.~\ref{fig:crossmodal_robustness} extends that observation to changing sensor reliability, showing that even adaptive fusion can fail if it is trained only under nominal visual conditions. An important finding is that the availability of sonar does not automatically improve perception. The sonar-context representation remains unaffected by the imposed optical degradation, yet fixed feature concatenation performs below the visual-only foundation-model baseline under $D_4$. Similarly, the adaptive gate trained only on clean observations performs well under moderate conditions but deteriorates sharply at the highest severity. These results demonstrate a distinction between \emph{sensor complementarity} and \emph{learned fusion robustness}.

Optical and acoustic sensing are naturally complementary in underwater environments. Optical imagery contains high-resolution appearance, texture, and color information but is highly susceptible to turbidity, illumination loss, scattering, and attenuation. Sonar is considerably less sensitive to optical visibility but generally provides lower spatial resolution and weaker appearance information. Consequently, neither sensing modality is uniformly superior across operating conditions \cite{wu2026multimodal}. Effective perception therefore requires a mechanism capable of exploiting this asymmetry rather than treating the modalities as equally reliable feature sources.

The fixed-fusion results reinforce observations reported in previous multimodal underwater studies. In the original UMOD experiments, simple feature addition and concatenation did not consistently improve detection and could introduce inter-modal interference, whereas explicitly designed cross-modal interaction produced substantially better results \cite{wu2026multimodal}. Similar motivations underlie adaptive visual-sonar fusion approaches developed for underwater recognition and detection \cite{abu2023multimodal,wang2023multimodal,chen2025uamfdet}. The present work extends this perspective from nominal multimodal accuracy to robustness under changing modality quality. Even when two modalities are informative in principle, a fusion function optimized for one reliability regime may fail when the statistical relationship between the modalities changes.

The poor $D_4$ performance of the clean-trained gate is particularly informative. Because this model already possesses a learnable mechanism for adjusting visual and sonar contributions, its failure cannot be explained simply by insufficient architectural flexibility. Instead, the training data provide little evidence from which the model can learn how the relationship between visual and acoustic information should change when optical reliability collapses. This result is consistent with multimodal robustness literature showing that explicit exposure to modality corruption or absence is often necessary for reliable adaptation \cite{neverova2016moddrop,ma2022smil}. The important implication is that adaptive fusion must be trained for adaptation; architectural adaptivity alone is not sufficient.

\subsection{Degradation-Aware Training Enables Reliability-Dependent Modality Use}
\label{sec:discussion_adaptive_fusion}

The strongest experimental result is obtained by degradation-aware gated fusion. Under $D_4$, the model achieves a balanced accuracy of $0.6152$, compared with $0.4610$ for the visual foundation-model baseline, corresponding to a $33.5\%$ relative improvement. The same model also substantially improves performance under $D_3$, reaching $0.6438$ compared with $0.4952$ for visual-only recognition. These improvements occur without modifying the frozen visual or sonar encoders. Instead, degradation-aware training acts only on the projection, fusion, and classification components.

The learned modality weights provide evidence that the improvement is associated with a systematic change in cross-modal reliance. Under clean conditions, the model is strongly vision dominated, assigning approximately $85.8\%$ of the average contribution to the visual representation and $14.2\%$ to sonar. Under $D_4$, the sonar contribution increases to approximately $41.3\%$. Thus, the model does not adopt a uniformly sonar-heavy strategy merely because acoustic information is available during training. It preserves visual dominance when optical observations remain informative and increases acoustic reliance only when severe degradation changes the available evidence.

This behavior is desirable for underwater multimodal systems because sensing reliability is inherently state dependent. A fixed assumption about modality quality is unlikely to remain valid across changes in depth, illumination, turbidity, target distance, and vehicle motion. The results therefore support reliability-conditioned fusion as a more appropriate design principle than constant multimodal weighting. Related observations have been made in multimodal autonomous perception, where adaptive cross-modal interaction improves robustness when one sensor deteriorates under adverse operating conditions \cite{wu2023robust}. The present results indicate that the same principle is particularly relevant for underwater optical-acoustic perception.

Importantly, the degradation level itself is not provided to the gating network. The model must infer the appropriate feature contribution from the modality representations. The increase in sonar weight at severe degradation therefore suggests that the projected visual representation contains sufficient evidence of reduced informativeness for the gate to alter its fusion behavior. This is preferable to a manually defined severity-dependent weighting schedule because real deployments will generally not provide an explicit ground-truth degradation label.

At the same time, the learned coefficients should not be interpreted as calibrated estimates of physical sensor reliability. They are task-dependent weights optimized to improve classification and may encode correlations involving scene context, object category, or representation geometry in addition to sensing quality. A more rigorous reliability interpretation would require explicit calibration or uncertainty modeling. Nevertheless, their systematic redistribution across $D_0$-$D_4$ provides useful diagnostic evidence that degradation-aware training changes how the model utilizes the two modalities.

\subsection{Fusion Benefits Depend on the Physical Nature of Degradation}
\label{sec:discussion_degradation_type}

The degradation-type ablation shows that acoustic assistance is not equally valuable for all forms of visual corruption. The largest relative gains occur under severe turbidity and blur, where degradation-aware fusion improves balanced accuracy by approximately $20.8\%$ and $15.4\%$, respectively. In contrast, isolated color attenuation at $D_4$ produces no measurable improvement over the visual-only foundation-model representation.

This distinction is physically meaningful. Turbidity and scattering reduce contrast and obscure object boundaries through the interaction of attenuation and veiling light, while blur directly suppresses high-frequency spatial structure. Both processes degrade geometric information that is important for distinguishing underwater objects. Sonar, in contrast, forms observations from acoustic returns and can preserve structural cues under conditions in which optical visibility is severely compromised. As a result, acoustic context provides genuinely complementary evidence in these cases.

Color attenuation produces a different type of distribution shift. Underwater propagation causes wavelength-dependent absorption, with longer wavelengths generally disappearing more rapidly with increasing path length \cite{akkaynak2018revised,akkaynak2019seathru}. However, a representation learned through large-scale self-supervision may remain relatively invariant to substantial changes in channel statistics as long as object shape and spatial structure remain sufficiently visible. This interpretation is consistent with the strong visual-only performance observed for the isolated color degradation. The absence of additional fusion benefit in that condition therefore does not imply that sonar is uninformative; rather, it indicates that the visual representation already contains adequate discriminative information for the evaluated objects.

These results also emphasize why underwater degradation should not be represented by a single generic corruption. Surveys of underwater enhancement and restoration have documented that illumination, scattering, attenuation, contrast reduction, and blur arise from different physical processes and produce different consequences for downstream vision \cite{li2020diving,shuang2024algorithms}. Evaluating robustness only under one synthetic corruption could therefore lead to incomplete conclusions regarding multimodal benefit. The present ablation suggests that the value of an auxiliary sensor depends strongly on which visual information has been lost.

\subsection{Implications for Underwater Robotic Perception}
\label{sec:discussion_robotics}

From a system-design perspective, the results suggest a hierarchical approach to robust underwater perception. Under favorable optical conditions, a strong visual representation may remain the primary information source because it provides richer appearance information than sonar. Under moderate degradation, foundation-model representations may retain sufficient invariance to avoid unnecessary dependence on the acoustic modality. As conditions become severe, however, the system should progressively exploit sensing channels that remain physically informative.

Such behavior is particularly relevant to remotely operated and autonomous underwater vehicles, where environmental conditions can vary substantially within a single mission. A vehicle may transition from relatively clear water to regions affected by suspended particles, sediment disturbance, artificial-light limitations, or rapid motion. A fusion system optimized only for nominal imagery may therefore exhibit unpredictable behavior precisely when redundancy is most important. Training with controlled variation in modality reliability provides a simple mechanism for encouraging more appropriate responses to these transitions.

The findings also have implications for computationally constrained marine platforms. The proposed approach does not require end-to-end fine-tuning of the foundation model. Instead, frozen modality encoders are combined with lightweight projection and gating components. This reduces the number of trainable parameters and makes the method conceptually compatible with resource-constrained deployment, where continuously fine-tuning or executing multiple large multimodal models may be infeasible. Although computational deployment is not explicitly evaluated in the present experiments, lightweight adaptive fusion provides a promising direction for combining foundation-model features with conventional sensor encoders.

More broadly, the results indicate that multimodal perception systems should be evaluated not only by their peak accuracy under nominal conditions, but also by their response to changing sensor reliability. A model that produces a small improvement on clean data but preserves substantially more performance when a primary modality deteriorates may be more valuable for autonomous operation than a model optimized exclusively for nominal benchmark accuracy. This perspective complements recent underwater multimodal detection work, which has primarily emphasized cross-modal alignment and feature-fusion accuracy \cite{wu2026multimodal,chen2025uamfdet}, by introducing reliability shift as an additional evaluation dimension.

\subsection{Limitations}
\label{sec:discussion_limitations}

Several limitations should be considered when interpreting the reported results. First, the study uses a relatively small synchronized subset of UMOD and retains five of the nine original target categories for the primary recognition experiments. Although grouped partitioning is used to reduce sequence-level leakage and all methods are evaluated on identical test objects, the limited number of independent test examples increases sensitivity to individual samples. Small non-monotonic changes, such as the increase in visual balanced accuracy between selected degradation levels, should therefore not be interpreted as evidence that degradation improves perception. A larger evaluation set and repeated sequence-level splits would provide tighter estimates of robustness and statistical variability.

Second, the multimodal experiments formulate the task as object-level recognition using ground-truth visual crops rather than end-to-end multimodal detection. This design intentionally isolates representation and fusion robustness from localization failure, but it also simplifies the perception problem. In a practical underwater detector, severe degradation may simultaneously affect object localization, classification, and cross-modal association. The dramatic degradation observed in the visual YOLO11n experiment demonstrates that localization itself becomes a major failure source under poor visibility. Consequently, the recognition results should be interpreted as evidence regarding cross-modal feature utilization rather than as direct estimates of end-to-end detection performance.

Third, the sonar representation is derived from the complete synchronized sonar frame rather than an explicitly aligned sonar object region. Optical and forward-looking sonar observations have substantially different image geometries, and direct transfer of optical bounding boxes to the sonar image is not valid without geometric calibration or learned correspondence \cite{wu2026multimodal}. The current formulation therefore evaluates acoustic \emph{context} rather than precise object-level sonar features. Some of the improvement may arise from scene-level correlations contained in the synchronized acoustic frame. Future work should incorporate explicit cross-modal correspondence, learned spatial alignment, or attention-based association to determine how much additional benefit can be obtained from target-specific acoustic information.

Fourth, only the visual modality is degraded. This provides a controlled experimental design in which the sonar branch acts as a stable complementary source, but real underwater environments may degrade both modalities simultaneously. Sonar can be affected by reverberation, multipath propagation, acoustic noise, geometric distortion, and resolution limitations \cite{yang2024sonar,wen2024sonar,wu2026multimodal}. A complete multimodal robustness benchmark should therefore include independent and joint corruption of optical and acoustic sensing, as well as missing-modality conditions.

Fifth, the current degradation functions are controlled approximations of underwater image deterioration rather than complete simulations of underwater radiative transfer. Real underwater imagery depends on depth, water type, illumination spectrum, camera response, target distance, and spatially varying scattering conditions. Physically based underwater image-formation models provide a more complete description of these effects \cite{akkaynak2018revised,akkaynak2019seathru}. Future benchmarks could use measured water parameters or physically calibrated degradation models to improve correspondence between synthetic reliability shifts and field conditions.

Finally, the present fusion method is intentionally lightweight. It does not perform token-level cross-attention, geometric correspondence learning, multi-scale fusion, or end-to-end adaptation of the encoders. More sophisticated architectures may achieve higher nominal performance or stronger robustness. The purpose of the current formulation is instead to isolate whether explicit training across modality-reliability changes is beneficial. The results indicate that this training principle is important even for a simple fusion mechanism and could therefore be incorporated into more advanced multimodal architectures.

\subsection{Future Directions}
\label{sec:discussion_future}

The findings motivate several extensions. A natural next step is to integrate degradation-aware reliability modeling into an end-to-end visual-sonar detector. Cross-modal fusion modules that address geometric misalignment, such as the attention-based mechanisms developed for UMOD, could be augmented with reliability-conditioned weighting so that both spatial correspondence and changing sensor quality are modeled jointly \cite{wu2026multimodal}. Such a system would permit direct evaluation using detection metrics under the full degradation sequence.

A second direction is uncertainty-aware fusion. Instead of learning only deterministic modality coefficients, each encoder could estimate predictive or representation uncertainty and expose this information to the fusion mechanism. The resulting system could distinguish between cases in which a modality is merely unusual and cases in which its evidence is genuinely unreliable. This would also enable the fusion weights to be evaluated against calibrated uncertainty rather than interpreted only diagnostically.

Third, robustness training should be extended to bidirectional and missing-modality settings. Visual degradation, sonar corruption, communication loss, and temporary sensor failure could be sampled independently during training. Strategies related to modality dropout and severely missing-modality learning \cite{neverova2016moddrop,ma2022smil} provide useful foundations for such experiments. An underwater perception system trained under these conditions could learn not only to increase sonar reliance when vision deteriorates, but also to revert toward optical evidence when acoustic measurements become unreliable.

Finally, evaluation on larger datasets and real field deployments is necessary to determine whether the learned reliability adaptation transfers beyond the controlled UMOD setting. Of particular interest is whether the increase in acoustic reliance observed under synthetic turbidity and blur also appears naturally as visibility changes during an underwater mission. Demonstrating such behavior would provide stronger evidence that degradation-aware multimodal fusion can support robust perception for autonomous marine robots operating under dynamic environmental conditions.

Overall, the results indicate that robustness in underwater multimodal perception is fundamentally a \emph{reliability-adaptation} problem. Strong pretrained visual representations delay performance degradation, and complementary acoustic sensing provides information that remains useful when optical visibility decreases. However, neither property alone is sufficient. The most robust behavior emerges when the fusion mechanism is explicitly exposed to changing modality quality and learns to redistribute its use of visual and acoustic evidence accordingly.

\section{Conclusion}
\label{sec:conclusion}

This work investigated robust underwater cross-modal perception under progressively degraded visual conditions, with particular emphasis on how multimodal fusion should respond when the reliability of the optical modality changes. Using a synchronized subset of UMOD, we introduced a controlled five-level degradation benchmark spanning illumination loss, wavelength-dependent attenuation, turbidity/scattering, and blur, and evaluated conventional visual detection, frozen DINOv2 representations, sonar context, fixed fusion, clean-trained gating, and degradation-aware gating. The results show that visual perception degrades substantially as optical quality deteriorates, while pretrained foundation-model representations retain a larger fraction of their clean-condition performance. However, simply introducing sonar does not guarantee robustness: fixed fusion and a gate trained only under nominal conditions become ineffective under severe degradation. In contrast, degradation-aware fusion preserves a balanced accuracy of $0.6152$ at $D_4$, compared with $0.4610$ for the visual foundation-model baseline, corresponding to a $33.5\%$ relative improvement. This improvement is accompanied by a substantial redistribution of learned modality contributions, with the mean sonar weight increasing from $14.2\%$ under clean conditions to $41.3\%$ under extreme degradation. The degradation-specific analysis further shows that acoustic assistance is most beneficial when visual structure is strongly compromised, particularly under severe turbidity and blur. Collectively, these findings support the central conclusion of this study: robust underwater multimodal perception depends not only on the availability of complementary sensing modalities, but also on explicitly training the fusion mechanism to operate across changes in their relative reliability.

Several limitations constrain the scope of the present conclusions. The experiments use a relatively small synchronized subset of UMOD and retain five target categories for the primary recognition task, which limits statistical power and increases sensitivity to individual test examples. The cross-modal experiments are formulated as object-level recognition using ground-truth visual crops rather than end-to-end multimodal detection, and the sonar branch represents the complete synchronized acoustic frame rather than a geometrically aligned target region. In addition, the degradation benchmark applies controlled synthetic perturbations only to the visual modality; it does not reproduce the full complexity of underwater radiative transfer or account for corruption of sonar measurements. The learned modality weights should therefore be interpreted as task-dependent fusion coefficients rather than calibrated estimates of physical sensor reliability. These design choices are intentional in this preliminary study because they isolate the effect of changing visual reliability, but they also mean that the reported gains should be viewed as evidence for the underlying reliability-adaptation principle rather than as a complete solution to end-to-end underwater multimodal perception.

Future work should extend this reliability-aware formulation toward full multimodal robotic perception. A natural next step is to integrate degradation-aware weighting into an end-to-end visual-sonar detector so that localization, cross-modal correspondence, and reliability adaptation are learned jointly. Such a model could combine the present degradation-aware mechanism with feature-alignment or attention-based fusion strategies designed specifically for the geometric heterogeneity of optical and forward-looking sonar sensing. Broader robustness studies should additionally consider independent and joint corruption of both modalities, missing-sensor conditions, uncertainty-aware fusion, repeated sequence-level evaluation, and physically calibrated degradation models. Most importantly, evaluation on the full UMOD benchmark and under real field conditions is needed to determine whether the observed increase in acoustic reliance transfers from controlled synthetic degradation to naturally varying visibility during underwater missions. If this behavior generalizes, degradation-aware cross-modal fusion could provide a practical foundation for underwater robots that maintain reliable perception by dynamically redistributing sensing reliance as environmental conditions change.
\bibliographystyle{elsarticle-harv} 
\bibliography{example_fixed}

\end{document}